\documentclass[runningheads,a4paper]{llncs}
\usepackage{amsmath}
\usepackage{amssymb}
\usepackage{subfigure}
\usepackage{graphicx}
\usepackage{multirow}
\usepackage{multicol}
\usepackage{color}

\usepackage{url}
\newcommand\norm[1]{\left\lVert#1\right\rVert}

\urldef{\mailsa}\path|{a.b, c.d, e.f, g.h|
\urldef{\mailsb}\path|i.j, k.l, m.n, o.p,|
\urldef{\mailsc}\path|emails}@urv.cat|    
\newcommand{\keywords}[1]{\par\addvspace\baselineskip
\noindent\keywordname\enspace\ignorespaces#1}

\begin{document}

\mainmatter  

\title{Conditional Generative Adversarial and Convolutional Networks for X-ray Breast Mass Segmentation and Shape Classification}

\titlerunning{Lecture Notes in Computer Science: Authors' Instructions}

%
%

\author{Vivek Kumar Singh\inst{1,\thanks{Corresponding Author: vivekkumar.singh@urv.cat}} \and Santiago Romani\inst{1} \and Hatem A. Rashwan\inst{1}  \and Farhan Akram\inst{2} \and Nidhi Pandey\inst{3,4} \and Md. Mostafa Kamal Sarker\inst{1} \and Saddam Abdulwahab\inst{1}\and Jordina Torrents-Barrena\inst{1} \and Adel Saleh\inst{1} \and Miguel Arquez\inst{4}\and Meritxell Arenas\inst{4}\and
Domenec Puig\inst{1}}

\authorrunning{V.K. Singh et al.}
%
\institute{DEIM, Universitat Rovira i Virgili, Spain. \and
Imaging Informatics Division, Bioinformatics Institute, Singapore.
\and Kayakalp Hospital, 110084 New Delhi, India. \and
Hospital Universitari Sant Joan de Reus, Spain.\\
}




%
%

\maketitle

\begin{abstract}
This paper proposes a novel approach based on conditional Generative Adversarial Networks (cGAN) for breast mass segmentation in mammography. We hypothesized that the cGAN structure is well-suited to accurately outline the mass area, especially when the training data is limited. The generative network learns intrinsic features of tumors while the adversarial network enforces segmentations to be similar to the ground truth. Experiments performed on dozens of malignant tumors extracted from the public DDSM dataset and from our in-house private dataset confirm our hypothesis with very high Dice coefficient and Jaccard index ($>94\%$ and $>89\%$, respectively) outperforming the scores obtained by other state-of-the-art approaches. Furthermore, in order to detect  portray significant morphological features of the segmented tumor, a specific Convolutional Neural Network (CNN) have also been designed for classifying the segmented tumor areas into four types (irregular, lobular, oval and round), which provides an overall accuracy about 72\% with the DDSM dataset.
\end{abstract}

\section{Introduction}

Mammography screening is the most reliable method for early detection of breast carcinomas~\cite{ChengSMHCD06}. Among diverse types of breast abnormalities, such as micro-calcifications or architectural distortion, breast masses are the most important findings since they may be pointing out the presence of malignant tumors \cite{kopans1998}. However, to locate masses and discern mass borders are difficult tasks because of their high variability, low contrast and high similarity with the surrounding healthy tissue, as well as their low signal-to-noise ratio \cite{elmore2009}.

Therefore, Computer-Aided Diagnosis (CAD) systems are highly recommended for helping radiologists in detecting masses, outlining their borders (mass segmentation) and suggesting their morphological features, such as shape type (irregular, lobular, oval and round) and margin type (circumscribed, obscured, ill-defined, spiculated).
Recent studies point out some loose correlations between mass features and molecular subtypes, i.e., Luminal-A, Luminal-B, HER-2 (Human Epidermal growth factor receptor 2) and Basal-like (triple negative), which are key for prescribing the best oncological treatment \cite{cho2016,liu2016,tamaki2011}.

Although it is impossible for an expert radiologist to discern the molecular subtypes from the mammography. Recently, a Convolutional Neural Network (CNN) was used to classify molecular subtypes using texture based descriptors of image crops of mass area \cite{Singh17CCIA}, which yielded an overall accuracy of $67\%$. 

In this paper, we present a novel approach for 1) breast mass segmentation based on conditional Generative Adversarial Networks (cGAN) \cite{isola2017image}, 2) to predict the mass shape type (irregular, lobular, oval and round) from the binary mask of the mass area. Beside these two contributions, this paper provides a study of the correlation between the mass shape and molecular subtypes.

\section{Related Work}

Numerous methods have been proposed to solve the problem of breast mass segmentation from a classical point of view, including techniques based on thresholding, iterative pixel classification, region growing, region clustering, edge detection, template matching and stochastic relaxation \cite{ChengSMHCD06,oliver2010review}.  

For the segmentation problem, some proposals rely on classic statistical models, such as structured Support Vector Machines, using Deep Belief Network or CNN features as their potential functions \cite{dhungel2015deep}. On the other hand, it is also possible to perform image segmentation based on the Fully Convolutional Network (FCN) approach \cite{long2015fully}. However, the classical FCN pipeline does not accurately preserve the objects boundaries. To overcome this drawback, an FCN network has been concatenated with a CRF layer taking into account the pixel position to enforce the compactness of the output segmentation \cite{zhu2016adversarial}.

In \cite{yang2017automatic}, a conditional Generative Adversarial Network (cGAN) has been used to segment the human liver in 3D CT images. However, this architecture is based on 3D filters, thus it is not suitable for mammography segmentation.

\section{Proposed Model}

\subsection{System overview}
\begin{figure}[!h]
\centering
\includegraphics[width=1.0\textwidth, height=0.5\textwidth]{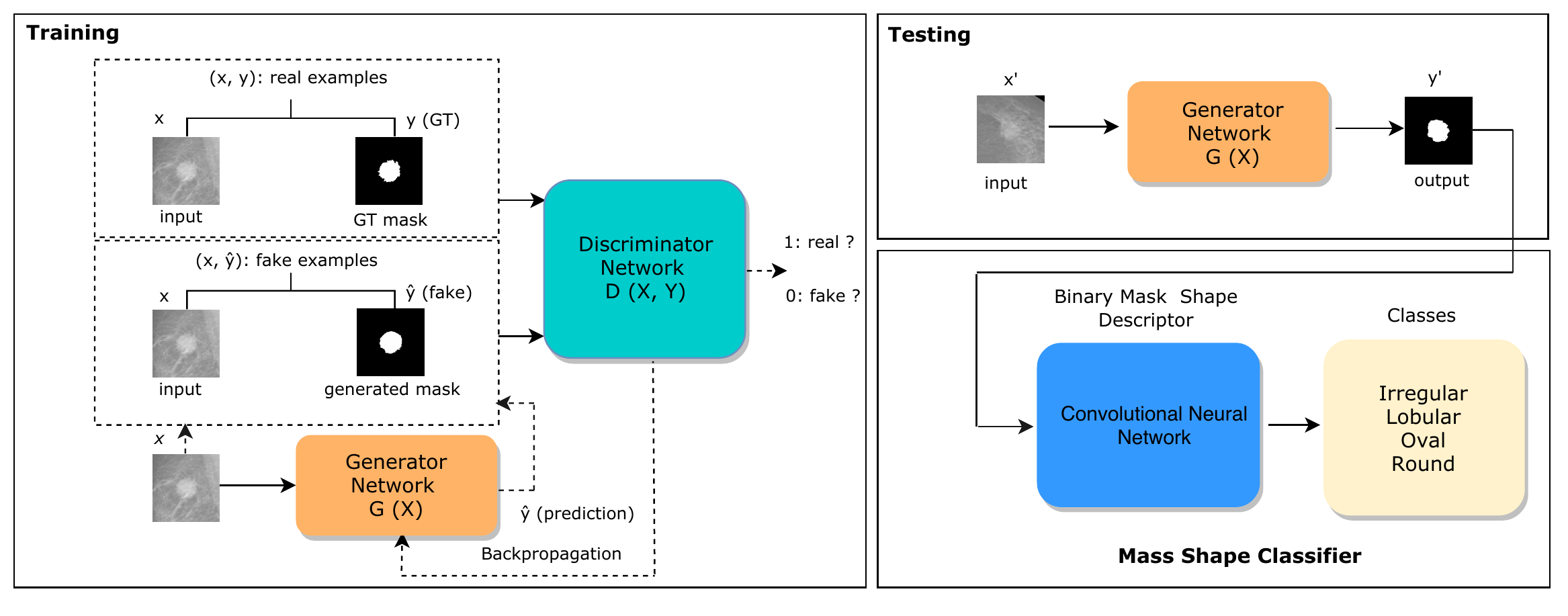}
\caption{Proposed framework for breast mass segmentation and shape classification}
\label{fig1:model}
\end{figure}

Fig.~\ref{fig1:model} represents the training phase of the proposed cGAN network for mass segmentation (left) as well as the full predicting workflow (right), defined by two stages. The first stage uses the generator part of the trained cGAN to automatically obtain a binary mask that selects the pixels (in white) that are supposed to correspond to the area of the breast mass, while ignores the pixels (in black) corresponding to healthy tissue. The input image is a squared crop of the mammogram containing the mass ROI. The input is reshaped to $256\times 256$ pixels size and the value of each pixel is scaled into a [0,1] range. For noise removal, we have regularized the image with Gaussian filter of 0.5 standard deviation. The second stage of the workflow uses a regular CNN trained to classify the obtained binary mask into one out of four classes of mass shape, which are irregular, lobular, oval and round.

\subsection{Mass segmentation model (with cGAN)}

We hypothesized that the cGAN structure proposed in \cite{isola2017image} would be perfect for segmentation, mainly for two reasons:

\begin{enumerate}
  \item The Generator network of the cGAN is an FCN network composed of two networks: encoders and decoders. Encoders can learn the intrinsic features of the masses and normal breast parenchyma (gray-level, texture, gradients, edges, shape, etc.), in turn decoders can learn how to mark up the binary mask according to the input features of the two output classes (mass/normal).
  \item The Discriminative network of the cGAN compares the generated binary mask with the corresponding ground truth to make them as similar as possible. Therefore, including the adversarial score in the loss computation of the generator strengthens its capabilities to provide a valid segmentation.
\end{enumerate}

This combination of generator/discriminator networks allows robust learning with very few training samples. Since both generative and discriminative networks are conditioned by observing the input image, thus the resulting segmentation is a “function” over the input pixels. Otherwise, regular GAN (unconditional) will infer the segmentation just from random noise, which obviously will not bind the mass appearance gathered by the x-ray with the output binary mask.

Let $x$ represents a mass ROI image, $y$ the corresponding ground truth segmentation, $z$ a random variable, $G(x, z)$ is the predicted mask, $\norm{ y-G(x, z)}_{1}$ is the L1 normalized distance between ground truth and predicted masks, $\lambda$ is an empirical weighting factor and $D(x, G(x, z))$ is the output score of the discriminator, the generator loss is defined as:

\begin{equation}
\ell_{Gen} (G,D) = E_{x,y,z} \big(-log (D(x,G(x,z)))\big)+ \lambda E_{x,y,z} \big(\norm{y-G(x,z)}_{1}\big), 
\end{equation}

As pointed out in \cite{isola2017image}, if we only use the L1 term, the obtained binary masks will be blurred since the distance metric averages all pixel differences. Therefore, including the adversarial term allows the generator to learn how to transform input images at fine-grained details (high frequencies), which results in sharp and realistic binary masks.

On the other hand, the L1 term is also necessary to boost the learning process, which otherwise may be too slow because the adversarial loss term may not properly formulate the gradient towards the expected mask shape.
The loss computation of the discriminator network is defined as:

\begin{equation}
\small
\ell_{Dis} (G,D) = E_{x,y} \big(-log (D(x,y))\big)+ E_{x,y,z} \big(-log (1-D(x,G(x,z)))\big),
\end{equation}

Hence, the optimizer will fit the discriminator network in order to maximize the real mask predication (by minimizing $-log (D(x, y))$ and to minimize the generated masks predication (by minimizing $-log (1-D(x, G(x, z)))$.

\subsection{Shape classification model (with CNN) }

For this stage, we have chosen a CNN approach instead of other classical approaches of extracting shape features (e.g. HOG, shape context) mainly because of the recent success of Deep Neural Networks in object recognition and segmentation tasks \cite{litjens2017survey}. Nevertheless, the input images for this stage (binary masks) do not render complex distribution of pixel values, just morphological structure, hence we hypothesized that a rather simple CNN (i.e., two convolutional layers plus two fully connected layers) will be sufficient to learn a generalization of the four mass shapes.

\section{Experiments}

To evaluate the performance of the proposed models, two datasets have been used: Digital Database for Screening Mammography (DDSM) \cite{heath2000digital} and our private in-house dataset of mammograms obtained from Hospital Universitari Sant Joan de Reus-Spain. For numerical assessment of the performance of the proposed mass segmentation, we have computed Accuracy, Dice Coefficient, Jaccard index (i.e., Intersection over Union (IoU)), Sensitivity and Specificity~\cite{vacavant2012benchmark}. 

\subsection{Datasets}
\textbf{DDSM dataset:} It is a publicly available database including about 2500 benign and malignant breast tumor masses, with ground truths of different shape classes. From malignant cases, we have selected 567 mammography images (330, 108, 90 and 39 images of irregular, lobular, oval and round shapes, respectively). We have used this dataset for training both segmentation and shape classification models.

\textbf{Reus hospital dataset:} It contains 194 malignant masses distributed into four molecular subtypes of breast cancer: 64 Luminal-A, 59 Luminal-B, 34 Her-2 and 37 Basal-like. This dataset is used to test the segmentation model and to make an analysis between shape mass and molecular subtype distributions.

\begin{table}[!b]
\centering
\caption{Accuracy, Dice coefficient, Jaccard index, Sensitivity and Specificity from the two architectures of cGAN (Auto-Encoder and Unet), FCN, U-Net and CRFCNN evaluated on DDSM and our private dataset. Best results are marked in bold.}
\label{Table1}
\begin{tabular}{|c|c|c|c|c|c|c|}
\hline
Dataset                      & Methods         & Accuracy        & Dice            & Jaccard         & Senstivity      & Specificity     \\ \hline
\multirow{5}{*}{DDSM}        & FCN             & 0.9114          & 0.8480          & 0.7361          & 0.8193          & 0.9511          \\ \cline{2-7}
                             & U-Net           & 0.9308          & 0.8635          & 0.7896          & 0.8365          & 0.9552          \\ \cline{2-7}
                             & CRFCNN          & 0.8245          & 0.8457          & 0.7925          & 0.8421          & 0.8975          \\ \cline{2-7}
                             & cGAN-AutoEnc     & 0.9469          & 0.9061          & 0.8283          & 0.8975          & 0.9666          \\ \cline{2-7}
                             & cGAN-Unet        & \textbf{0.9716} & \textbf{0.9443} & \textbf{0.8944} & \textbf{0.9274} & \textbf{0.9871} \\ \hline
\multirow{5}{*}{Private}     & FCN             & 0.9484          & 0.8698          & 0.7799          & 0.8002          & \textbf{0.9905} \\ \cline{2-7}
                             & U-Net           & 0.8647          & 0.7442          & 0.6622          & 0.6921          & 0.8641          \\ \cline{2-7} 
                             & CRFCNN          & 0.7542          & 0.6135          & 0.5247          & 0.7126          & 0.7458          \\ \cline{2-7}
                             & cGAN-AutoEnc     & 0.9481          & \textbf{0.8894} & \textbf{0.8008} & \textbf{0.9726} & 0.9414          \\ \cline{2-7}
                             & cGAN-Unet        & \textbf{0.9555} & 0.8648          & 0.7618          & 0.8576          & 0.9750          \\ \hline

\end{tabular}
\end{table}

\subsection{Experimental results}
For the first stage, we have trained two versions of the proposed cGAN architecture, Auto-Encoder (i.e., without skip connections) and U-Net (i.e., with skip connections), and compared them with three models: FCN \cite{long2015fully}, U-Net \cite{ronneberger2015u} and CRFCNN \cite{dhungel2015deep} retrained for our data. For all experiments, the DDSM dataset is divided into training, validation and testing by 70\%, 15\% and 15\%, respectively. In turn, whole in-house private dataset samples are used for testing (see Table~\ref{Table1}). After segmentation, we have applied a post-processing morphological filtering (i.e., erosion and dilation) to remove the artifacts and small white regions from the binary masks generated by all compared methods.

\begin{figure}[!h]
	\centering
	\includegraphics[width=0.9\textwidth,height=0.3\textheight]{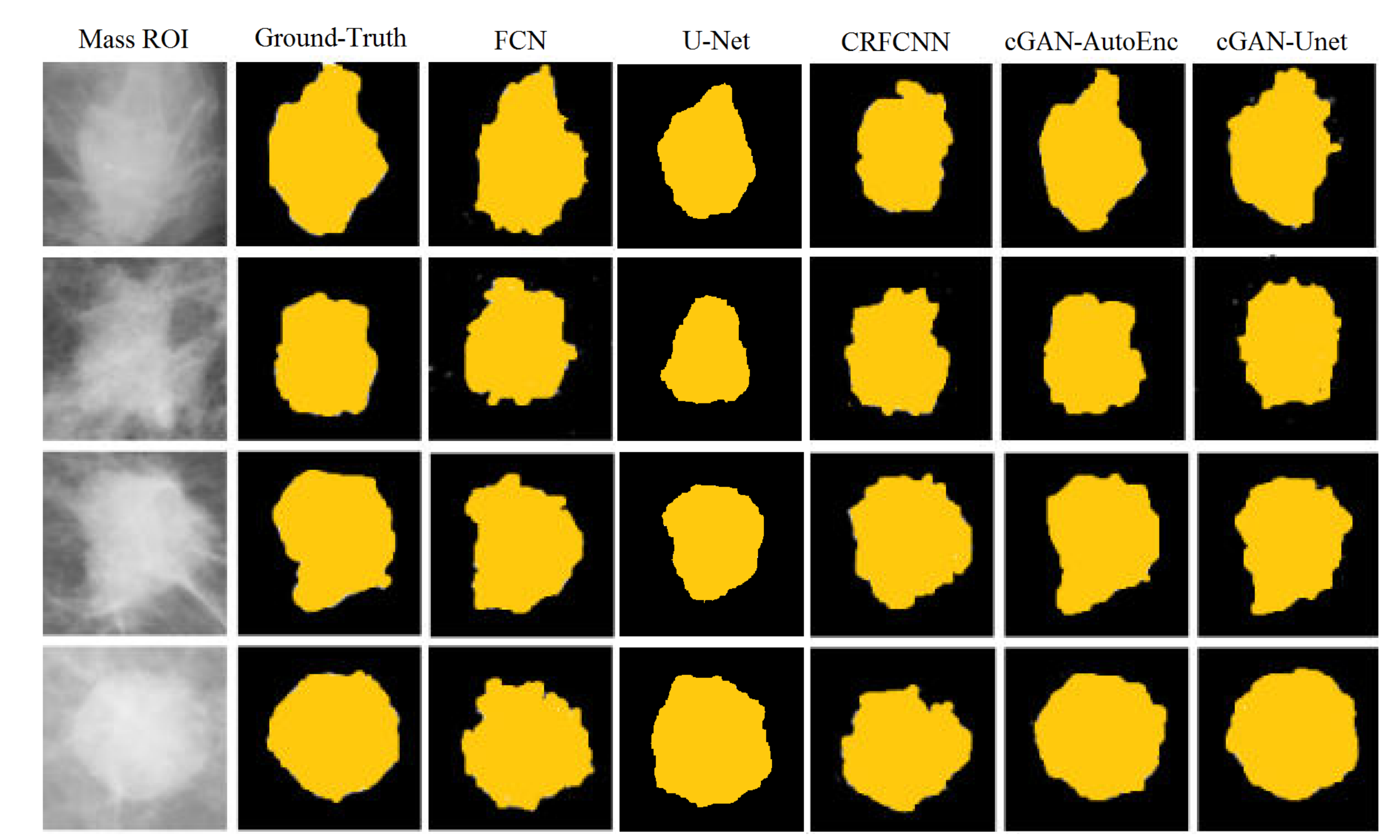}
	\caption{ Examples of hospital dataset mammographic mass ROI images (col 1), ground truth masks (col 2), and generated masks with FCN (col 3), CRFCNN (col 4), cGAN-AutoEnc (col 5), cGAN-Unet (col 6) and U-Net (col 7).}
	\label{fig5:Segmentation}
\end{figure}

The cGAN-Unet provides the best results of all computed metrics on the DDSM test samples, with very remarkable Accuracy, Dice and Jaccard scores (around 97\%, 94\% and 89\%, respectively). On the in-house private dataset, however, the cGAN-AutoEnc yields better results than the cGAN-Unet in terms of Dice, Jaccard and Sensitivity (+2\%, +4\% and +12\%, respectively), which indicates that the cGAN-AutoEnc has learned a more generalized representation of tumor features since it performs better on the dataset not used for training. Although the accuracy of cGAN-AutoEnc (94.81\%) is not higher than FCN (94.84\%) and cGAN-Unet (95.55\%), the former has obtained an impressive rate of true positives (97.26\%), which leads to the highest values of Dice and Jaccard (88.94\% and 80.08\%, respectively). The FCN model obtains the highest rate of true negatives (99.05\%) but its Sensitivity is poorer (80.02\%) than both cGAN versions, which indicates that it misses more real tumor area than the cGAN proposals. On the other hand, U-Net and CRFCNN provided even poorer results in both Sensitivity and Specificity for the private dataset, although the U-Net and FCN methods performed relatively well on the DDSM dataset. Some qualitative examples using our in-house private dataset are shown in Fig.~\ref{fig5:Segmentation}.

\vspace*{-3mm}

For training the second stage of shape classification, 80\% of the selected images from the DDSM dataset are used in training our classifier with their corresponding ground truth of mass shape labels, using a stratified 10 fold cross validation with 50 epochs per fold. The remaining 20\% of images are used for testing, obtaining an overall accuracy around 72\%.

Tumor shape could play an important role to predict the breast cancer molecular subtypes \cite{zhang2015identifying}. Thus, we have computed the correlation between breast cancer molecular subtypes classes of our in-house private dataset with the four shape classes. As shown in Table~\ref{Table2}, Luminal-A and -B groups are mostly assigned to irregular and lobular shape classes. In addition, some images related to Luminal-A are assigned to oval shape. In turn, oval and round masses give indications to the Her-2 and Basal-like groups, as well as some images related to Basal-like are moderately assigned to the lobular class.

\begin{table}[!h]
	\centering
	\caption{Distribution of breast cancer molecular subtypes samples from the hospital dataset with respect to its predicted mask shape.}
	\label{Table2}
	\begin{tabular}{|c|c|c|c|c|c|}
		\hline
		\begin{tabular}[c]{@{}c@{}}Shape classes/ \\ molecular subtypes\end{tabular} & Irregular & Lobular & Oval & Round & Total \\ \hline
		Luminal A                                                                    & 24        & 19      & 19   & 2     & 64    \\ \hline
		Luminal B                                                                    & 23        & 27      & 8    & 1     & 59    \\ \hline
		Her-2                                                                        & 7         & 3       & 10   & 14    & 34    \\ \hline
		Basal-like                                                                   & 2         & 13      & 4    & 18    & 37    \\ \hline
	\end{tabular}
\end{table}

\vspace*{-6mm}
\section{Conclusions}

In this paper, we propose two versions of cGAN networks for breast mass segmentation: cGAN-AutoEnc and cGAN-Unet. The generative network of both versions follows similar structures compared to FCN and U-Net networks, respectively. However, experimental results confirm that the inclusion of an adversarial network significantly improves the performance of the segmentation, about +6\% and +9\% in terms of Dice coefficient and Jaccard index, respectively on the public DDSM dataset. In turn, on our in-house private dataset, it yields an improvement of +2\% and +2\% with the two metrics. The CRFCNN provided worse test results in general. In addition, we have also proved that a rather simple CNN architecture is enough for distinguishing shape-related classes of the mass shapes from their binary masks. Future work aims to improve the overall accuracy (72\%) by using a large dataset and using a robust loss function, such as negative log likelihood and dice loss function for improving the convergence and accuracy of the proposed system.

\section*{Acknowledgement}
This research has been partly supported by the Spanish Government through project DPI2016-77415-R.

\bibliographystyle{splncs}
\bibliography{biblography} 

\end{document}